# SSI-GAN: Semi-Supervised Swin-Inspired Generative Adversarial Networks for Neuronal Spike Classification

Danial Sharifrazi[1*], Nouman Javed[1,2], Mojtaba Mohammadi[1], Seyede Sana Salehi[3], Roohallah Alizadehsani[1], Prasad N. Paradkar[2], U. Rajendra Acharya[4, 5], Asim Bhatti[1]

[1]*Institute for Intelligent Systems Research and Innovations (IISRI), Deakin University, Geelong, Australia*
[2]*CSIRO Health and Biosecurity, Australian Animal Health Laboratory, Geelong, Australia*
[3]*Department of Computer Engineering, Islamshahr Branch, Islamic Azad University, Islamshahr, Iran*
[4]*School of Mathematics, Physics and Computing, University of Southern Queensland, Springfield, Australia*
[5]*Center for Health Research, University of Southern Queensland, Springfield, Australia*

## Abstract

Mosquitos are the main transmissive agents of arboviral diseases. Manual classification of their neuronal spike patterns is very labor-intensive and expensive. Most available deep learning solutions require fully labeled spike datasets and highly preprocessed neuronal signals. This reduces the feasibility of mass adoption in actual field scenarios. To address the scarcity of labeled data problems, we propose a new Generative Adversarial Network (GAN) architecture that we call the **S**emi-supervised **S**win-**I**nspired **GAN** (SSI-GAN). The Swin-inspired, shifted-window discriminator, together with a transformer-based generator, is used to classify neuronal spike trains and, consequently, detect viral neurotropism. We use a multi-head self-attention model in a flat, window-based transformer discriminator that learns to capture sparser high-frequency spike features. Using just 1 to 3% labeled data, SSI-GAN was trained with more than 15 million spike samples collected at five-time post-infection and recording classification into Zika-infected, dengue- infected, or uninfected categories. Hyperparameters were optimized using the Bayesian Optuna framework, and performance for robustness was validated under fivefold Monte Carlo cross-validation. SSI-GAN reached 99.93% classification accuracy on the third day post-infection with only 3% labeled data. It maintained high accuracy across all stages of infection with just 1% supervision. This shows a 97–99% reduction in manual labeling effort relative to standard supervised approaches at the same performance level. The shifted-window transformer design proposed here beat all baselines by a wide margin and set new best marks in spike-based neuronal infection classification. This paper presents a robust, data-hungry approach to neurovirology, showing that transformer attention combined with semi-supervised adversarial learning can effectively detect subtle viral signatures in neuronal spike data with very little expert labeling.

[*] Corresponding author: D. Sharifrazi, Institute for Intelligent Systems Research and Innovations (IISRI), Deakin University, Geelong, Victoria 3216, Australia
Email Addresses: d.sharifraz@deakin.edu.au (D. Sharifrazi), n.javed@deakin.edu.au (N. Javed), m.mohammadikarachi@deakin.edu.au (M. Mohammadi), sanaa.salehi@iau.ir (S. Salehi), r.alizadehsani@deakin.edu.au (R. Alizadehsani), prasad.paradkar@csiro.au (P. Paradkar), Rajendra.Acharya@unisq.edu.au (R. Acharya), asim.bhatti@deakin.edu.au (A. Bhatti).

# 1. Introduction

Zika virus (ZIKV) and Dengue virus (DENV) belong to the Flaviviridae family and the Ortho flavivirus genus [1]. Other viruses of this family include West Nile (WNV), Japanese encephalitis (JEV), and Yellow fever virus (YFV) [1]. ZIKV was identified in a rhesus macaque in Uganda in 1947 [2]. It has had outbreaks in Africa, the Americas, Asia, and the Pacific region since 2007 and has become a serious global health concern over time [3]. This virus was recorded in more than 50 countries and infected around 390 million people worldwide as of now [4]. During these times, serious nervous system disorders, such as Guillain-Barré syndrome, encephalitis, myelitis, and neuralgia, and also developmental problems such as microcephaly and eye malformations, were largely observed [5].

ZIKV is mainly transmitted by *Aedes aegypti* mosquitoes [6]. It can also spread directly between humans through mother to fetus, sexual contact, and blood transfusion [7]. Studying how viruses affect mosquitos' behavior and their nervous system, as a part of the virus life cycle, will help us to understand the transmission of mosquito-borne diseases [8]. There are many evolutionary similarities in the nervous systems of mammals and insects [9]. Both of them use the same chemical messengers and receptors [10]. The similar basic structure of the receptors in both groups allows us to compare functions in their bodies [11]. Their nervous systems also have structural and functional similarities [12]. There is a central nervous system and a peripheral nervous system in both species [13]. There are some differences, but these similarities suggest that studying ZIKV and DENV neural activity in insects can help understand their pathogenesis and infection mechanisms in humans [14].

ZIKV's effects on the nervous system have been widely studied using different models, including in vitro studies with human pluripotent stem cell (hPSC)-derived neural progenitor cells and organoids, as well as in-vivo research in mouse models by means of different viral strains [15]. In a previous study, which focused on the mosquito life cycle, neural spikes produced by uninfected, ZIKV-infected, and DENV-infected *Aedes aegypti* were recorded by microelectrode array technology [16]. Analyzing the behavior and neural spikes showed virus neurotropism, which affects mosquito neurons and causes an increase in neural activity and behavioral changes [16]. These changes included different flight movements, more bites, more feeding, and effects on reproductive performance, such as fecundity and fertility [16]. These changes can make viral transmission easier by increasing mosquito-human interactions [16].

Neural spike classification is effective for analyzing the behavior of the neurological system among different methods, because it isolates spikes, the brain's signals for transmitting information [17]. Studying mosquito neuronal activity and neurons' spike classification is important for understanding the viral infection mechanisms of ZIKV and DENV [18]. Using deep learning to classify spikes in these insects can help us understand how these viruses affect both mosquitoes and humans [19]. Deep learning can make data processing automated and improve accuracy by reducing human error in classification tasks [20]. In addition, it can identify patterns and correlations in neuronal activity that may not be easily recognizable through traditional methods [21]. There are almost no studies that have focused on deep learning-based spike classification of insect data for ZIKV and DENV detection [18] using semi-supervised learning.

Review shows that several traditional methods have been used for spike classification [22]. For example, Lewicki used a Bayesian approach [23], while Letelier and Weber used wavelet analysis to classify neural spikes based on their time-frequency features, isolating signals from noise and overlapping activity [24]. Haas, Cohen et al. proposed a variance-based template matching model, which uses a CMOS variance estimator to classify data in real time [25]. Takekawa, Isomura et al. introduced a hybrid algorithm that combines multimodality-weighted principal component analysis and clustering techniques [26]. Traditional models perform well on certain benchmarks, but often exhibit low accuracy and poor generalization, and most of the time, they work effectively only on specific datasets [27]. On the other hand, deep learning models show impressive performance according to their ability to filter noise, enhance signal decoding, and produce more reliable results [28]. For example, Meyer, Schanze, et al.[28] developed a single-layer multilayer perceptron network of spike waveforms. Their results show that the model has a good performance, with accuracy between 89%-95% depending on the noise level [29]. Liu, Feng et al. proposed a method that uses convolutional neural networks (CNNs) and long short-term memory for spike classification [30]. Li, Tang et al. proposed a hybrid deep reinforcement learning- based method for spike classification on imbalanced data [31]. In their study, the problem is modeled as a Markov sequence decision and a dynamic reward function is used to increase the effect of the minor group in the data during training [31]. Saif-ur-Rehman, Ali et al. introduced 'SpikeDeep-Classifier,' an automatic spike-sorting algorithm that integrates supervised learning with unsupervised K-means clustering. The main idea is to apply the same set of hyperparameter settings to all datasets [32]. Park, Eom et al. used fully connected neural networks to classify neural spikes from extracellular recordings, which was effective in handling similar spike shapes and high noise levels [33]. They generated pseudo-labels using principal component analysis and clustering, which have been shown to be effective for handling unlabeled data [33]. García, Suárez et al. proposed a hybrid modular computational system for classifying spikes in multi-unit neural recordings without previous knowledge of waveform structure [34]. The proposed system consists of preprocessing and processing modules, and uses a hybrid unsupervised multilayer artificial neural network to identify features, cluster spikes, and filter noise [34]. Bernert and Yvert developed a spiking neural network using spike-timing-dependent plasticity for real-time, unsupervised spike sorting [35]. Their results showed high accuracy on noisy data, adaptability to different spike patterns, and suitability for multi-channel recordings [35].

As shown in the studies above, deep learning is highly effective for spike classification [36]. However, one major challenge when working with ZIKV and DENV data is labeling [36]. The process of labeling large datasets involves intensive lab work and relies significantly on skilled technicians to accurately classify complex biological data [37, 38]. This manual process is not only time-consuming and expensive but also unscalable as data volume increases [37, 38]. An efficient solution for this problem is a semi-supervised algorithm [37, 38]. Semi-supervised learning techniques use a few labeled samples as well as many unlabeled ones, which makes lab work easier, cuts costs, and improves model accuracy [37, 38]. The successful use case of semi-supervised algorithms can be seen in text sentiment analysis [39], speech recognition [40], and DNA sequencing [41]. In this regard, this paper uses semi-supervised techniques for data preparation and labeling.

Given the significance of analyzing neural activity in mosquitoes for investigating viral infections, this paper proposes a hybrid method that combines semi-supervised techniques and deep learning for neural spike classification in mosquitoes infected with ZIKV and DENV. Notably, state-of-the-art deep learning methods have been employed, and experimental lab-generated data that includes neural spike information from uninfected, DENV-infected, and ZIKV-infected *Aedes aegypti* mosquitoes are used to train the model. To the best of the authors' knowledge, this is the first study addressing the spike classification problem in insects using semi-supervised techniques and advanced deep learning models. By integrating semi-supervised techniques with deep learning, this study aims to fill existing gaps in neural spike classification research.

## 2. Data Collection

### 2.1. Virus Strains and Preparation

ZIKV was isolated from a Cambodian human in 2010 (GenBank KU955593). It was grown in *Aedes albopictus* C6/36 cells several times and harvested 72 hours post-infection. DENV2 was from an Australian soldier infected in East Timor and was also amplified seven times in C6/36 cell lines. L15 culture media was used to prepare these viruses, which were then used to infect mosquitoes and primary neuron cultures.

### 2.2. Mosquito Rearing and Infections

*Aedes aegypi* mosquitoes were cultured in a controlled laboratory condition at 26°C with 60–70% humidity, and a 14-hour light followed by a 10-hour dark modeling natural dawn and dusk. Infectious blood was given to female mosquitoes at day 7 for an hour at 37°C, and infected ones were selected for subsequent experiments.

### 2.3. Immunohistochemistry (IHC)

We used IHC to detect viral presence in head sections from 11 ZIKV-infected and 4 control mosquitoes using the 4G4 antibody (1:100 dilution). We performed 20-minute antigen retrieval at 97°C and blocked endogenous peroxidase. In the next step, signal amplification was performed using EnVision FLEX+ Mouse Linker and HRP-labeled secondary antibody with AEC chromogen substrate. The slides were then examined and photographed at 4× and 40× magnifications with a Leica microscope.

### 2.4. Primary Neuron Isolation

Three to four days after emergence, brains of mature female *Aedes aegypi* were removed and transferred into L15 culture solution containing tryptose phosphate broth (10%), fetal calf serum (10%), penicillin–streptomycin (50 U/mL), and fungizone. Between 15 and 20 brains were collected and mechanically fragmented in a 2 ml tube using gentle pipetting. After 2 centrifugations, the pellet was suspended in 50 µL of fresh medium for cell culture.

### 2.5. Neuron Culture and Infection

A similar method was used to prepare primary neuron cultures for confocal microscopy, TCID50 and RT-PCR according to standard protocols. For confocal assays, polyethyleneimine (PEI) was applied to the coverslips for 30 minutes at 37°C, washed three times with sterile water, and dried in a biosafety cabinet. 25 µl of the cell mixture was pipetted into each well of a 24-well plate. The plates were incubated at 28ºC for 30 minutes to enable cell adhesion, after which 1 ml of L15 medium with 10% FCS was added. On day 7, when neuron cultures had matured, cells were infected at a Multiplicity of Infection (MOI).

The microelectrode array (MEA) neuron culture protocol was similar, with modifications based on Hales et al. [42]. The modifications include using laminin after PEI for precoating, different cell densities for plating, and media changes every 3-4 days, except on the day before recording. Finally, a different volume of viruses was used for infection.

## 2.6. Data acquisition with microelectrode array

*A. Recording system*

Neuronal network activity was recorded using a multichannel system's MEA, with 60 planar electrodes arranged in a square grid and one electrode as the ground reference. For online recording of neuronal signals, the MEA60inv system was used, and action potentials were sampled at 30 kHz.

All recordings were performed in a Biosafty Cabinet II (BSCII) to avoid viral contamination, and at 28°C to optimize neuronal life and function. Recordings were done 5-10 minutes after MEA placement, allowing neurons to stabilize from mechanical stress. Natural neural activities were recorded at 0, 2, 3 and 7 days after infection for 30 minutes, and at 8 dpi for 15 minutes during Gabazine-induced tests. After each recording session, 50% of the culture medium was replaced. The initial voltage recordings were processed with a high-pass Butterworth filter to detect low-frequency noise.

*C. Recording timeline*

After plating mosquito neurons onto a MEA, cultures were allowed to form networks for 7 days in vitro. On day 7, neurons were infected with ZIKV and DENV (as a control). Each experiment included six recordings of neural signals. The first recording was on made day 7 prior to the infection, and it was used as a reference. Post-infection recordings were done on days 2, 3 and 7.

To perform continuous electrical recordings at a high sampling rate of 30 kHz, the brains of female mosquitoes were removed, and the neurons were cultivated on MEAs. The recordings were subsequently filtered with a high-pass filter to eliminate background noise.

The standard deviation (std) for each signal type, including Control, DENV2, and ZIKV, was determined on the first day of infection and was 6.1 µV, 4.08 µV, and 5.44 µV, respectively. Next, we use the rounded average of the standard deviations as the threshold. Only neuronal spikes that occurred when the signal crossed a threshold set at twice the noise level (10 µV) would be tallied.

## 2.7. Final Dataset

15,728,580 samples of data were collected from three groups, including ZIKV, DENV and control at different days after infection (0, 1, 2, 3 and 7). From each group, 1,048,572 recordings were made.

# 3. Proposed Methodology

## 3.1 Spike Detection

The first preprocessing step aims to clean the raw dataset and detect neural spikes. Spike detection begins by applying a high-pass Butterworth filter to the collected mosquito-neuron recordings, using a 700 Hz cut-off frequency [43] and a 30 kHz sampling rate [44, 45]. This suppresses low-frequency noise and preserves the high-frequency components characteristic of spike events. In the filtered signal, amplitudes greater than +10 µV or less than −10 µV are considered spike occurrences. All other values are set to zero, leaving only meaningful spike information while discarding baseline noise. Because the proposed model is trained on spike sequences rather than raw signals, this step ensures that the classifier focuses exclusively on informative neural activity. Next, the cleaned spike signal is segmented into non-overlapping 100-sample windows to form the input sequences for classification. This window length was determined experimentally: shorter sequences did not capture sufficient temporal structure, whiereas longer ones increased computation without improving performance. Because the segments are non-overlapping, no spike value appears in more than one sample.

### 3.2 Data Labelling

After spike detection and segmentation, the dataset is split into training and test sets, with 20% of the data randomly selected for testing. For semi-supervised learning, only 3% of the training data is randomly chosen and labelled, while the remaining 97% remains unlabeled. Additionally, 1% of the labelled subset is reserved as validation data to monitor model performance during training and mitigate the risk of overfitting. All selections, including training, validation, labelled, and unlabeled sets. are made randomly.

### 3.3 Generator Model

Following data preparation, the architectures of the generator and discriminator in the proposed **SSI-GAN** are constructed. Although many GAN studies rely on CNN-based generator and discriminator pairs, the strong performance of attention mechanisms in various fields motivates the use of transformer-based components in this work.

Two generator variants were examined: one using convolutional layers, and one incorporating attention blocks. Because the generator in semi-supervised GANs primarily serves to support discriminator learning rather than producing high-fidelity synthetic samples, the final performance of the classifier showed difference between the two designs. We will reveal more core architecture design hypotheses in the ablation studies section. Thus, both convolutional and transformer-style layers are used in the generator.

The generator receives a 128-dimensional random noise vector. It progressively transforms it into a synthetic $100 \times 60$ spike-like sequence through dense projection, reshaping, Conv2DTranspose layers, batch normalization, LeakyReLU activations, and final tanh output activation.

### 3.4 Discriminator Model (Novel Swin-Inspired Shifted-Window Design)

Unlike the generator, the discriminator serves as the **final classifier** and is therefore central to the performance of the proposed model. Since the input spike sequences are sparse, containing mostly zeros except at spike locations. It is essential for the model to focus selectively on informative temporal regions. This requirement naturally aligns with self-attention mechanisms.
In this study, we introduce a Swin-inspired, shifted-window transformer discriminator tailored for 1-dimensional neural spike sequences. Unlike the original hierarchical Swin Transformer, which performs multi-stage down-sampling, our discriminator:

- does not apply hierarchical merging, preserving the full 100 × 60 resolution of the spike sequence,
- segments the sequence into fixed-size windows,
- applies self-attention locally within each window, and
- uses shifted windows via a rolling operation to alternate attention boundaries between layers.

This simplified shifted-window design allows efficient local modeling while enabling cross-window interaction in alternating layers, which is critical for sparse spike patterns that may occur near window boundaries. The absence of hierarchical down-sampling makes the model especially suitable for neural data, where maintaining timing granularity is essential.

The resulting discriminator architecture contains:

- an input projection layer,
- two shifted-window Swin-inspired transformer blocks,
- global average pooling,
- dense layers with LeakyReLU and dropout, and
- an output softmax layer producing Control, Dengue, and Zika classifications.

Fake samples from the generator are also included during training to support semi-supervised learning. Figure 1 illustrates the full SSI-GAN architecture. This novel non-hierarchical Swin-inspired discriminator with shifted windows is a key contribution of the SSI-GAN framework.

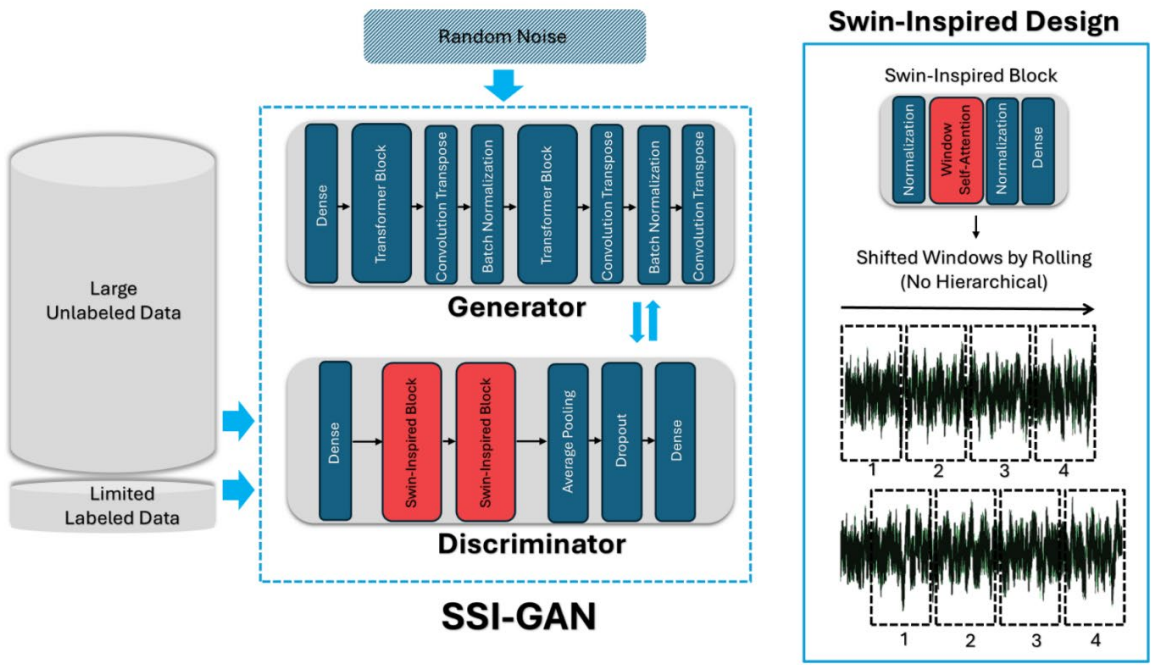

**Figure 1.** Proposed SSI-GAN model architecture based on the Swin-inspired discriminator.

## 3.5 Hyperparameter Optimization

Given the large number of hyperparameters across both the generator and discriminator components, a combined strategy is adopted. Critical hyperparameters are optimized using Bayesian methods via the Optuna platform [46], while secondary parameters are tuned manually based on prior literature [47]. After 20 Optuna iterations, the best configuration achieved an accuracy of 99.48%. Table 1 summarizes the search ranges and optimal values.

**Table 1.** Summary of model hyperparameter names, range, and best recommendation based on the Optuna platform.

| *Name of Hyperparameters* | *Range of Hyperparameters* | *Best Outcome* |
|---|---|---|
| **Dimension of Random Noise** | [64, 100, 128] | 128 |
| **Transformer Head Size** | [64, 128, 256] | 64 |
| **Number of Heads** | [2, 4, 8] | 4 |
| **Feed Forward Dimension** | [32, 64, 128] | 128 |
| **Number of Transformer Blocks** | [1, 2, 3, 4] | 2 |
| **Dropout Rate** | [0.1 – 0.5] | 0.29 |
| **Learning Rate** | [1e-5 – 1e-3] | 0.0009 |
| **Batch Size** | [64, 128] | 128 |

## 3.6 Model Training and Evaluation

Using the selected hyperparameters, the SSI-GAN is trained for 500 iterations. The generator and the proposed shifted-window Swin-inspired discriminator are jointly optimized throughout training. Monitoring the accuracy curves (Figure 2) confirms that joint training is stable and without overfitting.

To ensure statistical reliability, the full methodology is executed five times using Monte Carlo cross-validation [48, 49], where all data partitions (training, testing, validation, labeled, and unlabeled) are newly randomized in each run. Final results are reported as the mean and standard deviation across the five executions.

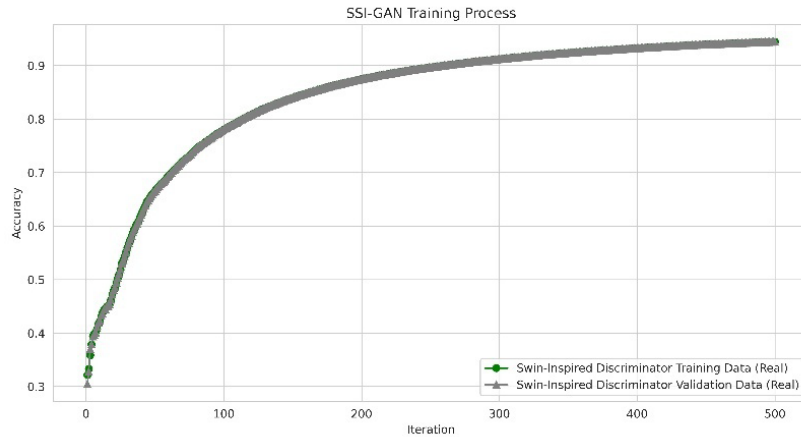

**Figure 2.** Training process of the generator and discriminator model after 500 iterations.

### 3.7 Method Summary

A high-level overview of the proposed method is shown in Figure 3, and the detailed implementation steps are provided in pseudocode format in Appendix A. In summary, the methodology consists of the following steps:

1. Record mosquito neural signals for Control, Dengue, and Zika conditions.

2. Apply high-pass filtering to detect spikes and remove non-spike values.

3. Segment spike signals into 100-sample non-overlapping windows.

4. Split the dataset into test, labeled (3%), validation (1% of labeled), and unlabeled sets.

5. Construct and optimize the SSI-GAN, featuring a novel shifted-window Swin-inspired discriminator.

6. Train the model using semi-supervised learning, where labeled, unlabeled, and synthetic data jointly improve classification.

7. Repeat the entire pipeline five times using Monte Carlo cross-validation. Figure 3 summarizes the complete SSI-GAN methodology.

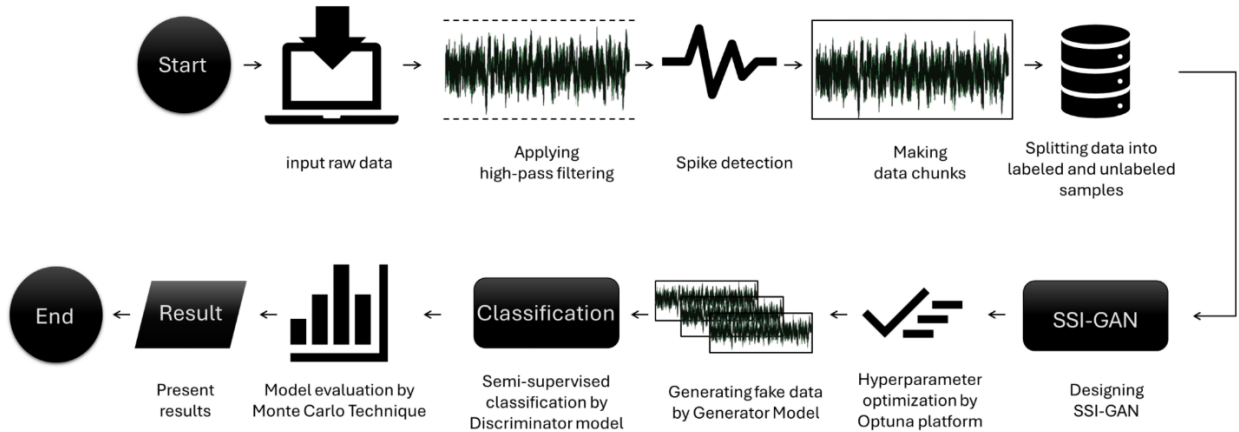

**Figure 3.** Proposed methodology steps in brief.

## 4. Experimental Results

### 4.1 Fully Supervised Architecture Outcomes

Our experimental dataset comprises 15,728,580 neural spike samples collected from *Aedes aegypti* mosquitoes. The dataset consists of three experimental groups: ZIKV-infected, DENV-infected, and uninfected controls. Neural recordings were obtained using a 60-MEA with 59 planar electrodes arranged in a square grid configuration. The action potential sampling rate was set to 10 kHz. Data samples were collected at multiple time points, including day 0 (pre-infection), day 1, day 2, day 3, and day 7 post-infection (dpi). Each group contributed 1,048,572 recordings per time point. We applied a high-pass Butterworth filter with a 700 Hz cutoff frequency at a 30,000 Hz sampling rate in the preprocessing pipeline, which helped to isolate neural spikes from background noise. Subsequently, signals with amplitudes exceeding ±10 standard deviations from baseline fluctuations were identified, after which, all non-spike values were set to zero. The continuous signal was then segmented into non-overlapping sequences of 100-time units. This segmentation was done empirically to balance the pattern recognition capability with computational efficiency.

Following a semi-supervised learning method, the dataset was partitioned by randomly selecting 20% of the total samples as the test set. From the remaining 80% of the training data. Additionally, 1% of the labeled subset was set aside as validation data to evaluate training progress and prevent overfitting. All data partitioning was performed by using Monte Carlo cross-validation with 5 random repeats.

We assessed the model using four standard classification metrics: accuracy, precision, recall, and F1-score. Results are reported as mean values across 5 Monte Carlo cross-validation runs. The comprehensive evaluation across five temporal stages of viral infection (0, 1-, 2-, 3-, and 7 days post-infection) reveals significant performance differences among baseline architectures. It highlights the superiority of the proposed SSI-GAN approach for neuronal spike classification. Table 2 presents the quantitative results for the baseline models and compares these models with our proposed method.

**Table 2.** Summary of performances (%) for various models obtained for the Optuna platform.

| dpi | Evaluation Metric | LSTM | CNN-LSTM | RNN | CNN-RNN | CNN | GRU | CNN-GRU | SSI-GAN |
|---|---|---|---|---|---|---|---|---|---|
| 0 dpi | AC | 33.20 | 86.11 | 50.25 | 57.67 | 79.14 | 85.48 | 91.57 | **99.45** |
|  | PR | 24.93 | 86.67 | 50.86 | 59.49 | 81.22 | 86.67 | 91.94 | **99.46** |
|  | RE | 33.74 | 86.12 | 50.20 | 57.70 | 79.17 | 85.47 | 91.59 | **99.46** |
|  | FS | 23.01 | 86.02 | 49.39 | 57.03 | 78.89 | 85.28 | 91.56 | **99.45** |
| 1 dpi | AC | 33.39 | 78.93 | 55.68 | 68.95 | 84.31 | 84.41 | 91.08 | **99.61** |
|  | PR | 15.43 | 80.26 | 56.56 | 69.85 | 84.77 | 85.12 | 91.78 | **99.62** |
|  | RE | 33.39 | 78.93 | 55.66 | 68.99 | 84.27 | 84.41 | 91.12 | **99.61** |
|  | FS | 19.25 | 78.99 | 55.46 | 68.88 | 84.35 | 84.49 | 91.11 | **99.62** |
| 2 dpi | AC | 34.03 | 77.09 | 53.41 | 66.40 | 74.26 | 75.77 | 83.21 | **97.09** |
|  | PR | 20.81 | 77.83 | 56.50 | 69.03 | 75.68 | 77.02 | 84.04 | **97.34** |
|  | RE | 33.94 | 77.07 | 53.43 | 66.38 | 74.23 | 75.72 | 83.18 | **97.07** |
|  | FS | 20.71 | 77.22 | 53.32 | 66.77 | 74.62 | 75.80 | 83.29 | **97.10** |
| 3 dpi | AC | 33.00 | 70.78 | 78.95 | 85.94 | 86.38 | 87.80 | 90.32 | **99.82** |
|  | PR | 11.00 | 71.77 | 79.30 | 86.24 | 87.46 | 88.29 | 90.60 | **99.82** |
|  | RE | 33.33 | 70.79 | 78.99 | 85.93 | 86.33 | 87.80 | 90.27 | **99.82** |
|  | FS | 16.54 | 70.41 | 78.98 | 85.97 | 86.01 | 87.69 | 90.17 | **99.82** |
| 7 dpi | AC | 33.14 | 69.36 | 54.82 | 66.70 | 76.98 | 82.67 | 84.93 | **99.93** |
|  | PR | 11.05 | 71.06 | 55.15 | 67.72 | 77.80 | 84.32 | 85.65 | **99.93** |
|  | RE | 33.33 | 69.45 | 54.83 | 66.73 | 77.05 | 82.63 | 84.91 | **99.93** |
|  | FS | 16.59 | 68.91 | 53.95 | 66.58 | 76.49 | 82.50 | 84.87 | **99.93** |

\* Abbreviations: AC: Accuracy, PR: Precision, RE: Recall, FS: F1-score.

Based on the results presented in Table 2, simple recurrent architectures consistently performed poorly across different experimental time points. Long Short-Term Memory (LSTM) networks achieved the lowest classification accuracies, ranging from 33.00% to 34.03% across all time points. This indicates that they barely exceeded random chance for three-class classification. The severe underperformance stems from the vanishing gradient problem inherent in training deep neural networks, such as standard LSTM architectures. Recurrent Neural Network (RNN) models exhibited marginal improvements, with accuracies ranging from 50.25% to 78.95% across infection stages. The mid-stage infection at 3 dpi (78.95%) produced spike patterns with sufficient event-onset interval information for basic recurrent processing to detect the infection effectively. However, RNNs struggled at early (1 dpi: 55.68%) and late (7 dpi: 54.82%) infection stages, indicating sensitivity to variable event-onset interval patterns.

The performance of hybrid CNN-RNN architectures ranged from 57.67% at 0 dpi to 85.94% at 3 dpi. This implies that combining convolutional feature extraction with recurrent temporal modeling yielded only moderate improvements. In this approach, convolutional layers effectively extracted local spike waveform features, while recurrent components captured temporal dependencies. However, a problem with this approach is that the sequential nature of RNN processing limits its capability to implement parallelization and reduces computational efficiency. Furthermore, pure CNN architectures demonstrated substantially stronger performance, with accuracies ranging from 74.26% to 86.38%. The best CNN performance was achieved at 3 dpi (86.38%), indicating that spatial feature learning through convolutional filters effectively captures distinguishable spike waveform characteristics during peak viral neurotropic effects. However, CNNs exhibited performance degradation at 7 dpi (76.98%), which suggests prolonged infection introduces waveform variability that challenges purely convolutional approaches.

CNN-LSTM hybrid models produced inconsistent results, with accuracies ranging from 69.36% to 86.11%. This finding highlights the complexity of hybrid architecture design for spike classification tasks.

Gated recurrent units (GRU) achieved consistent performance improvements over RNNs, with accuracies ranging from 75.77% to 87.80%. GRUs maintained stable performance across infection stages, with peak accuracy at 3 dpi (87.80%). The gating mechanisms also effectively addressed gradient vanishing problems as observed in LSTMs while maintaining computational efficiency.

CNN-GRU hybrid architectures have emerged as the most substantial traditional supervised baseline, achieving accuracies ranging from 83.21% and 91.57%. This architecture consistently outperformed all other non-GAN baselines by combining CNN spatial feature extraction with GRU temporal processing and gating mechanisms. The peak performance of 91.08% at 1 dpi indicates the effectiveness of this approach, particularly during early infection days, when spike patterns exhibit emerging viral signatures.

Most architectures achieved moderate to high accuracies reflecting neuronal variability among experimental groups. Performance generally declined at 1 dpi. This suggests that viral infection has begun to alteri neuronal dynamics without yet fully developing discriminative signatures. The critical phase of viral neurotropism occurs at 2-3 dpi, when infection-induced neurophysiological changes reach maximum distinction; therefore, most architectures demonstrate performance recovery at this stage. However, performance declined slightly at 7 dpi for many architectures, suggesting that chronic infection may introduce greater spike pattern variability or stabilization.

Our SSI-GAN consistently outperformed all baseline architectures across all temporal stages and labeling ratios. With only 3% labeled data, the proposed model achieved accuracies of 99.45%, 99.61%, 97.09%, 99.82%, and 99.93% at 0, 1, 2, 3, and 7 dpi, respectively. Transformer architectures' self-attention ability, with a shifted window approach inspired by the Swin Transformer model, enabled the capture of long-range temporal dependencies in spike sequences, compared to convolutional or recurrent approaches. Multi-head attention layers process different aspects of spike patterns and identify subtle discriminative features that traditional architectures usually overlook. Moreover, the semi-supervised learning framework effectively leverages large amounts of unlabeled spike data through adversarial training. This feature reduces the dependency on expensive, labor-intensive, and exhaustive labeling stages while maintaining high classification accuracy. A comparison of the proposed SSI-GAN model with different labeled and unlabeled portions is illustrated in the next section.

### 4.2 Labeled and Unlabeled Data Variation Outcomes

Figure 4 compares the SSI-GAN model using varying amounts of labeled training data across all metrics and the average of five dpis. Three distinct labeling scenarios were evaluated: 1% labeled data (99% unlabeled), 2% labeled data (98% unlabeled), and finally, 3% labeled data (97% unlabeled).

Significant performance improvements were noted at 2 and 3 percent labeled data, but the proposed model performed similarly to the standard GAN with only 1 percent labeled data. The accuracy, precision, recall, and F1-score values of the SSI-GAN were above 97 % with 2% labeled data. Outperforming 1% labeled data by roughly 1.34% in accuracy and by comparable margins across other metrics, the best performance was achieved with 3% labeled data, at which all four metrics exceeded 99%.

Even with less supervision (1% labeled data), the proposed model maintained competitive performance relative to the fully supervised methods mentioned in Section 4.1, achieving an accuracy of over 96% across the five time points. These findings demonstrate that robust classification is possible even with sparse labeled data, particularly when utilizing sophisticated SSI-GAN architectures. For electrophysiological analysis, where labeled data acquisition is expensive and time-consuming, this highlights the usefulness of semi-supervised learning frameworks.

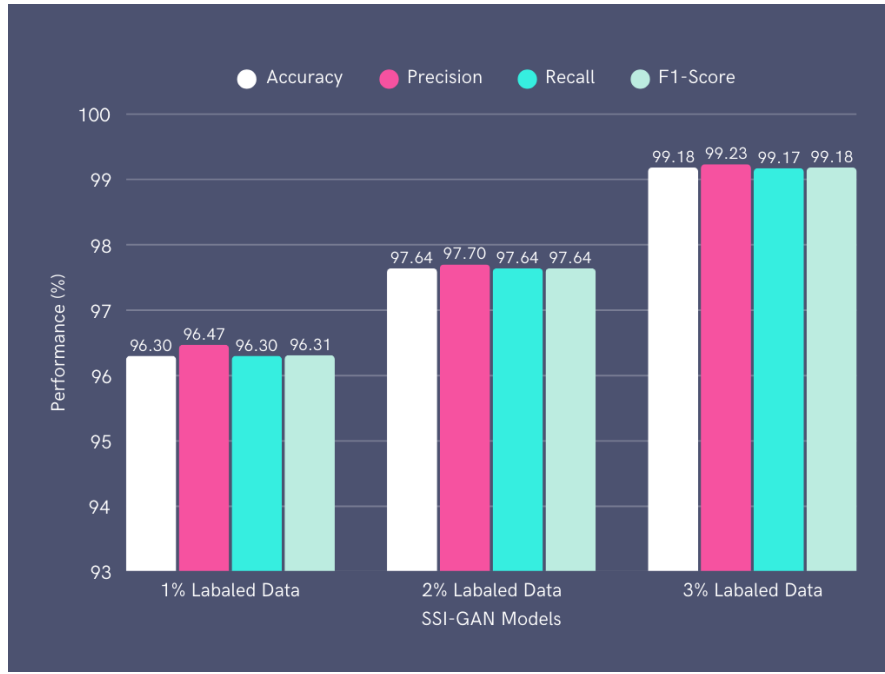

**Figure 4.** Comparison of the proposed SSI-GAN model with different amounts of labeled and unlabeled data.

## 4.3 Overall Outcomes Over 5-Fold Cross Validation

Table 3 shows the overall classification performance of the proposed SSI-GAN model with a three percent labeled data across all infection time points (0, 1-, 2-, 3-, and 7-days post-infection). The findings are presented as mean values with standard deviations, derived from five separate Monte Carlo cross-validation runs with different random initializations.

Standard deviations are reported to show statistical robustness, whether seeds are chosen at random or a specific labeled sample is allocated. Low standard deviations (typically less than 0.05 percent) across all configurations attest to the model's consistent ability to achieve high classification accuracy with little labeled supervision, which is essential for real-world implementation in neuroscience applications where manual spike labeling requires substantial resources.

**Table 3.** Overall performance (%) with standard deviations for 5 runs of Monte Carlo.

| *dpi* | *Accuracy* | *Precision* | *Recall* | *F1-Score* |
|---|---|---|---|---|
| *0* | 99.06 ± 0.46 | 99.08 ± 0.44 | 99.06 ± 0.46 | 99.06 ± 0.46 |
| *1* | 97.49 ± 03.89 | 97.60 ± 03.68 | 97.51 ± 03.86 | 97.50 ± 03.87 |
| *2* | 97.86 ± 0.93 | 97.95 ± 0.85 | 97.86 ± 0.95 | 97.86 ± 0.94 |
| *3* | 99.20 ± 0.76 | 99.20 ± 0.76 | 99.20 ± 0.77 | 99.20 ± 0.76 |
| *7* | 99.84 ± 0.11 | 99.84 ± 0.11 | 99.84 ± 0.11 | 99.84 ± 0.11 |
| *Overall Performance* | **98.69** | **98.73** | **98.69** | **98.69** |

## 5. Ablation Studies

To understand the contribution of each architectural component in the proposed SSI-GAN, we conducted an ablation study by systematically varying the generator and discriminator architectures across five alternative baselines. These configurations were evaluated across different days post-infection (dpi), and the performance was measured using standard classification metrics: Accuracy (AC), Precision (PR), Recall (RE), and F1-score (FS). The full results are presented in Table 4.

The first key observation is that Baseline 1, which uses both a transformer-based generator and a Swin Transformer discriminator, performs very poorly (e.g., 32–34% accuracy across all dpis). This suggests that using the Swin Transformer discriminator falls into overfitting. Because the whole architecture of Swin Transformer is dramatically complex for our case study, the discriminator leads to overfitting. Note that we used only the Swin Transformer architecture as the discriminator, not the pre-trained Swin Transformer proposed by Liu et al. [50] since the pre-trained one, trained on image data, which is irrelevant to our case study.

Baseline 2, which pairs a CNN generator with a plain transformer discriminator, performs much better, reaching over 97% accuracy at 0 dpi and even higher at other levels. This shows that self-attention in the discriminator has a stronger impact on classification than in the generator.

Baseline 3, with both generator and discriminator using standard transformers, performs comparably but slightly underperforms the proposed baseline. This indicates that adding attention in the generator does not necessarily improve performance and may introduce unnecessary complexity.

Interestingly, Baseline 4, which uses a CNN generator and our Swin-inspired discriminator, achieves stronger results than the previous baselines, particularly at mid-to-high dpi levels (e.g., 99.28% at 1 dpi and 99.18% at 3 dpi). This suggests that the Swin-inspired architecture contributes significantly to the model's robustness, especially when using real-world noisy data.

Baseline 5, a fully CNN-based GAN, performs worse than all other setups except Baseline 1, with particularly low scores at 3 and 7 dpi. This confirms that convolutional models alone are insufficient for this task, likely due to their inability to focus on sparse, spike-dense regions of the input.

In contrast, the proposed SSI-GAN, which combines a transformer-based generator with a Swin-inspired discriminator using a shifted-window mechanism (without hierarchical down-sampling), consistently outperforms all baselines across all dpis. It achieves near-perfect classification performance, with accuracy exceeding 99% at all levels and reaching 99.93% at 7 dpi.

Ablation Highlights:

- The discriminator architecture has a greater impact on performance than the generator.

- Our Swin-inspired discriminator with shifted windows (not hierarchical) enables effective local and cross-window feature learning, particularly well-suited for sparse spike data.

- Adding attention in the generator is not essential, but combining it with our Swin-inspired discriminator leads to the most stable and accurate results.

- The full SSI-GAN model outperforms all ablations, confirming that both components in our final architecture contribute to state-of-the-art classification on neural spike data.

**Table 4.** Different core design ablation based on our proposed algorithm. Baseline 1: Generator: Transformer, Discriminator: Swin Transformer. Baseline 2: Generator: CNN, Discriminator: Transformer. Baseline 3: Generator: Transformer, Discriminator: Transformer. Baseline 4: Generator: CNN, Discriminator: Swin-Inspired. Baseline 5: Generator: CNN, Discriminator: CNN. Proposed SSI-GAN: Generator: Transformer, Discriminator: Swin-Inspired.

| Dpi | Evaluation Metric (%) | Baseline 1 | Baseline 2 | Baseline 3 | Baseline 4 | Baseline 5 | SSI-GAN |
|---|---|---|---|---|---|---|---|
| 0 dpi | AC | 32.96 | 97.15 | 98.45 | 93.97 | 96.21 | **99.45** |
| | PR | 10.98 | 97.36 | 98.47 | 94.11 | 96.42 | **99.46** |
| | RE | 33.33 | 97.12 | 98.46 | 93.97 | 96.21 | **99.46** |
| | FS | 16.52 | 97.16 | 98.46 | 94.00 | 96.18 | **99.45** |
| 1 dpi | AC | 33.76 | 97.82 | 97.32 | 99.28 | 95.38 | **99.61** |
| | PR | 11.25 | 97.88 | 97.38 | 99.28 | 95.84 | **99.62** |
| | RE | 33.33 | 97.83 | 97.32 | 99.28 | 95.41 | **99.61** |
| | FS | 16.82 | 97.83 | 97.32 | 99.28 | 95.35 | **99.62** |
| 2 dpi | AC | 34.04 | 87.93 | 90.36 | 98.48 | 96.87 | **97.09** |
| | PR | 11.34 | 89.35 | 90.38 | 98.48 | 97.03 | **97.34** |
| | RE | 33.33 | 87.88 | 90.30 | 98.48 | 96.86 | **97.07** |
| | FS | 16.93 | 87.85 | 90.32 | 98.48 | 96.85 | **97.10** |
| 3 dpi | AC | 32.10 | 98.71 | 98.55 | 99.18 | 98.53 | **99.82** |
| | PR | 14.24 | 98.71 | 98.55 | 99.20 | 98.54 | **99.82** |
| | RE | 32.54 | 98.68 | 98.55 | 99.19 | 98.53 | **99.82** |
| | FS | 17.31 | 98.69 | 98.55 | 99.19 | 98.53 | **99.82** |
| 7 dpi | AC | 33.65 | 99.12 | 98.09 | 82.75 | 95.56 | **99.93** |
| | PR | 11.21 | 99.13 | 98.15 | 83.73 | 96.73 | **99.93** |
| | RE | 33.33 | 99.11 | 98.06 | 82.69 | 95.56 | **99.93** |
| | FS | 16.78 | 99.12 | 98.07 | 82.36 | 95.46 | **99.93** |

# 6. Discussion

This study introduces the first application of semi-supervised learning for insect neural spike classification as an indicator of viral infection, thereby opening new research avenues at the intersection of computational neuroscience, vector biology, and machine learning. Fully supervised learning frameworks in other studies depend on fully labeled training data. This study therefore addresses a key limitation identified in existing research on neuronal spike classification. A thorough review of existing work reveals that all previous deep learning efforts on spike classification have employed fully supervised paradigms, which necessitate exhaustive manual annotation of spike waveforms. The newly proposed SSI-GAN achieves state-of-the-art performance (99.93% peak accuracy) using just 3% labeled data.

The key benefit of our semi-supervised framework is that it cuts down the need for manual labeling by 97-99% when put side by side with old fully supervised methods. Manual spike sorting done by expert neuroscientists stands as a big block in electrophysiology studies because of the long hours or days used up for each recording session on visual checks, waveform groupings and classifying proof. This method is becoming increasingly costly and cannot be scaled up.

Beyond data efficiency, the transformer architecture, combined with the shifted-window approach, provides temporal pattern recognition capabilities that the state-of-the-art lacks. It is actually the ability of the self-attention mechanism to model long-range dependencies across the whole sequence of spikes (100-time units) that explains such huge improvements in performance: 12.27-24.40% higher accuracy than CNN models and an improvement over LSTM networks by 65.68-67.98%.

This becomes extremely important for sparse, heavily preprocessed spike data where most values are set to zero and relevant discriminative features may be temporally distant. Several other advantages include the use of a multi-head attention design that enables parallel processing of multiple temporal relationships and the parallel capture of complementary features, which distinguish ZIKV, DENV, and control neuronal activity. Model generalization across five temporal stages (0, 1, 2, 3, 7 dpi) maintains high accuracy ranging from 97.09%-99.93%. This temporal robustness stands in stark contrast to baseline models exhibiting steep performance declines at transitional infection stages, with the CNN-based GAN showing a 2.83% accuracy drop at 1 dpi and traditional models declining by 10-15%.

Monte Carlo cross-validation with different random seeds is an easy way to keep checks and reproducibility. Most people do not talk about it in machine learning. Low standard deviations across all metrics confirm that reported performance gains reflect the strong learning capacity rather than artifacts of favorable data splits or specific labeled-sample selection.

Table 5 presents a comparative performance analysis with further studies. The scarcity of machine learning research on this specific dataset (ZIKV/DENV) reflects the developing state of computational approaches to insect neuronal electrophysiology in vector-borne disease contexts.

Table 5. Comparative performance of spike classification methods on neuronal data.

| Study | Method | Learning Paradigm | Labeled Data Required | Peak Accuracy(%) | Dataset |
|---|---|---|---|---|---|
| **Gaburro et al., 2018 [16]** | NeuroSigX (manual classification tool) | Manual | 100% | 100 | ZIKV/DENV *Aedes aegypti* neurons |
| **Sharifrazi et al., 2024 [18]** | CNN + XGBoost | Fully Supervised | 100% | 97.99 | ZIKV/DENV *Aedes aegypti* neurons (60-channel MEA) |
| **Meyer et al., 2019 [29]** | MLP | Fully Supervised | 100% | 89-95 | Mammalian cortical spikes |
| **Liu et al., 2020 [30]** | CNN-LSTM | Fully Supervised | 100% | 91-93 | Benchmark spike datasets |
| **Park et al., 2020 [33]** | FCN | Fully Supervised | 100% | 87-90 | Multi-electrode arrays |
| **Proposed Study, 2026** | **SSI-GAN** | **Semi-Supervised** | **1-3%** | **98.69** | **ZIKV/DENV *Aedes aegypti* neurons** |

As shown in Table 5, all previous deep learning approaches are fully supervised, which makes them very expensive in terms of manual labeling, and therefore, not suitable for high-throughput vector-borne disease research, and the state-of-the-art performance achieved by the proposed semi-supervised framework with 1-3% labeled data demonstrates a fundamental methodological advance beyond incremental accuracy improvements, which can be practically deployed in neuroscience applications that have limited resources. By comparing the same data, this direct comparison can eliminate confounding variables arising from differences in dataset characteristics, recording quality, or biological variability, thereby providing the strongest evidence for the superiority of the proposed method. Compared with the larger spike classification literature using different datasets, the proposed approach still holds significant advantages. Other fully supervised deep learning approaches, such as the multilayer perceptron networks of Meyer (with 89-95% accuracy) and the CNN-LSTM hybrid of Liu (with 91%-93% accuracy), also perform well in spike classification.

This study represents a significant methodological advance, but it has some limitations:

- Origin and environmental control: The dataset was only obtained from *Aedes aegypti* primary neurons cultured in vitro in controlled laboratory conditions (26 °C, 60–70 percent humidity), which may not fully reflect the biological diversity observed in wild mosquito populations exposed to varying ecological and environmental stressors.

- Species and viral scope: The analysis was restricted to one mosquito species (*Aedes aegypti*) and two flavivirus strains (ZIKV and DENV2), so the results may not apply to other *Aedes* species or to other neurotropic flaviviruses (e.g., West Nile, Japanese encephalitis, or yellow fever viruses).

- Computational demands: The transformer-based model is capable, but it requires significantly more computational resources to train than the simple baseline architectures, which can be a practical limitation.

- Static evaluation setting: It was only evaluated on static datasets, not in real-time deployment scenarios where latency, streaming data, and continuous inference are essential.

- Interpretability challenges: While the attention mechanism helps to weight features, the features driving classification decisions may not be readily interpretable, which can complicate interpretability and biological understanding.

## 7. Conclusion

This study presents the first application of Swin-inspired semi-supervised generative adversarial networks (SSI-GAN) to neuronal spike classification in virus-infected mosquitoes. Our proposed model achieved peak classification accuracy of 99.93% at 3 days post-infection using only 3% labeled training data. This approach exceeded the further GAN baselines and all traditional deep learning architectures evaluated in this study. The transformer discriminator's self-attention mechanism, powered by a shifted window approach, further demonstrated superiority over common attention-based, convolutional, and recurrent approaches, with accuracy improvements ranging from 12.27% to 24.40% over pure CNN models and from 65.68% to 67.98% over LSTM networks across temporal stages.

The semi-supervised learning framework addresses a critical limitation in neuroscience research by reducing manual labeling requirements by approximately 97-99% while maintaining competitive or superior accuracy compared to fully supervised methods. The ability to achieve 95.04-97.83% accuracy across infection time points, even with minimal supervision (1% labeled data), demonstrates the practicality for high-throughput electrophysiological screening applications. The temporal performance patterns revealed through systematic evaluation at 0, 1-, 2-, 3-, and 7-day post-infection illuminate the biological progression of ZIKV and DENV neurotropism. The peak classification accuracy coincided with maximal viral-induced neurophysiological changes at 3 dpi.

This work provides a quantitative framework for understanding how flaviviruses alter mosquito neural dynamics, with implications for manipulating vector behavior, transmission ecology, and intervention development.

For future work, extending the approach to additional mosquito vector species and to comprehensive flavivirus panels would establish the approach's generalizability. Investigating even lower labeled-data ratios (such as zero-shot mechanisms) through active learning or self-supervised pre-training could further reduce the requirements for manual labeling. This approach enabled completely unsupervised spike classification and the development of real-time classification systems deployable for live neuronal recordings, which could be used in closed-loop experiments to dynamically alter experimental parameters based on automated detection of the infection state. Integration with concurrent behavioral tracking would enable correlation of neuronal activity patterns with mosquito locomotion, blood-feeding propensity, and host-seeking behaviors. Lastly, explainable AI approaches, such as attention weight visualization and gradient-based saliency mapping, can help identify which temporal positions and spike characteristics are most influential on the classification decision.

**Acknowledgment**
We sincerely thank Dr. Julie Gaburro for her invaluable efforts and dedication in collecting and preparing the dataset used in this study. Her contributions were essential to the success of this research.

**Availability of Data and Materials**
The dataset used in this study can be accessed by contacting the corresponding author, subject to ethical approval from the Australian Centre for Disease Preparedness (CSIRO), a statutory authority of the Australian Government, which holds the legal rights to the data. The code for the proposed method is publicly available on the GitHub platform via the provided link.

**Conflict of Interest**
The authors declare that there is no conflict of interest regarding the publication of this paper.

**Funding Statement**
The paper was funded by Deakin University for open access publication.

**Ethics Statement**
Animal ethics and consent to participate declarations are not applicable for this research as it did not involve humans or animals.

# Appendix A

The complete procedure of the implementation of the proposed SSI-GAN architecture for the detection of dengue, Zika and control classes using pseudocode.

---

**Proposed Algorithm:** Semi-supervised Swin-Inspired Generative Adversarial Network (SSI-GAN)
**Input:** Raw Multielectrode Array Recording Signals
**Output:** Detected Classes (Control, Dengue, and Zika) with 3% Labelled Data

---

Start

1. Spike Data Preparation
1.1 Read raw signals from 3 classes:
- CT (label = 0), DN (label = 1), ZK (label = 2)
- Source: read_each_class_all(dpi)
1.2 Segment raw signals into segments:
- Length = 100 samples per segment
- Use sliding segmentation (non-overlapping)
1.3 Concatenate and assign class labels
1.4 Apply high-pass Butterworth filter:
- Cutoff = 700 Hz, Sampling rate = 30 kHz
- Removes low-frequency noise
1.5 Extract spikes by thresholding:
- Zero out values in (−10, 10) μV range
- Retain high-amplitude events as spikes
Output: spike_data $\in \mathbb{R}^{N \times 100}$, labels

2. Generator $G(z)$
Input: Latent vector $z \in \mathbb{R}^{128}$
- Project and reshape to 2D feature map
- Apply 2 transformer blocks (attention over spatial layout)
- Upsample using Conv2DTranspose layers
- Output: synthetic spike signal $\in \mathbb{R}^{100 \times 60}$

3. Swin-Inspired Discriminator $D(x)$
Input: Sequence $x \in \mathbb{R}^{100 \times 60}$
- Project input to embedding dimension
- Apply stacked Swin-Inspired Transformer1D blocks (windowed attention)
- Global average pooling → Dense → Softmax (3-class)
- For unsupervised: apply sigmoid on Softmax logits (real/fake)

4. GAN Architecture
- $G(z) \rightarrow D(x)$
- Discriminator used in both supervised and unsupervised branches
- Unsupervised branch outputs real/fake probability via custom sigmoid

5. Training Procedure
For t = 1 to T:
- Sample 3% labeled data for supervised training
- Train D on real (labeled) data and fake (generated) data
- Train G to maximize D's misclassification (via GAN loss)
- Log loss/accuracy every N iterations
Use:
- Label smoothing (real = 0.9)
- Dropout = 0.29

End